\documentclass{svproc}
\usepackage[utf8]{inputenc}

\usepackage{amsmath,amsfonts,bm}

\def\eqref#1{equation~\ref{#1}}

\def\1{\bm{1}}

\DeclareMathAlphabet{\mathsfit}{\encodingdefault}{\sfdefault}{m}{sl}
\SetMathAlphabet{\mathsfit}{bold}{\encodingdefault}{\sfdefault}{bx}{n}

\usepackage{algorithm}
\usepackage{algpseudocode}
\usepackage{amsmath}
\usepackage{graphicx}
\usepackage{listings}
\usepackage{xcolor}
\usepackage{url}
\usepackage{multirow}
\usepackage{subcaption}
\usepackage[T1]{fontenc}
\usepackage{inconsolata}
\usepackage{hyperref}

\begin{document}
\mainmatter

\title{PhaseNAS: Language-Model Driven Architecture Search with Dynamic Phase Adaptation}

\titlerunning{PhaseNAS}  

\author{Fei Kong \and Xiaohan Shan \and Yanwei Hu \and Jianmin Li}

\authorrunning{Fei Kong et al.}

\institute{
Qiyuan Lab, Beijing, China \\
\email{\{kongfei,shanxiaohan,huyanwei,lijianmin\}@qiyuanlab.com}\\
}

\maketitle

\begin{abstract}
Neural Architecture Search (NAS) is challenged by the trade-off between search space exploration and efficiency, especially for complex tasks. While recent LLM-based NAS methods have shown promise, they often suffer from static search strategies and ambiguous architecture representations.
We propose \textbf{PhaseNAS}, an LLM-based NAS framework with dynamic phase transitions guided by real-time score thresholds and a structured architecture template language for consistent code generation.
On the NAS-Bench-Macro benchmark, PhaseNAS consistently discovers architectures with higher accuracy and better rank. For image classification (CIFAR-10/100), PhaseNAS reduces search time by up to 86\% while maintaining or improving accuracy. In object detection, it automatically produces YOLOv8 variants with higher mAP and lower resource cost.
These results demonstrate that PhaseNAS enables efficient, adaptive, and generalizable NAS across diverse vision tasks.

\end{abstract}

\section{Introduction}
Neural Architecture Search (NAS) has become a foundational technique for automating neural network design in modern AI systems \cite{elsken2019nas} \cite{ying2019bench} \cite{real2019regula}. Its relevance is especially prominent in autonomous and intelligent systems, where task-specific adaptation and resource efficiency are critical. However, traditional methods such as evolutionary algorithms \cite{real2017large} and reinforcement learning \cite{zoph2016neural} face substantial computational costs, particularly in large-scale model scenarios \cite{lin2021zen}. Gradient-based NAS approaches like DARTS \cite{liu2018darts} and meta-learning strategies have been proposed to improve efficiency, yet scalability and generalization remain open challenges \cite{wistuba2019survey}.

Recent advances explore integrating large language models (LLMs) into NAS, enabling LLMs to reason over architecture design using natural language \cite{zheng2023gpt4nas} \cite{nasir2024llmatic} \cite{rahman2024lemo}. While these approaches show promise, they face a fundamental resource allocation problem: different search phases require different computational capabilities, yet current methods use static LLM configurations throughout the entire search process. This leads to substantial inefficiency: using large LLMs for broad exploration wastes computational resources, while small LLMs are insufficient for fine-grained architectural refinement. Current LLM-based NAS methods suffer from several key limitations stemming from this resource mismatch:

(1) \emph{Resource Mismatch in Search Phases}: Current frameworks use fixed LLM configurations regardless of search phase requirements. Smaller models can efficiently handle broad exploration, while architectural refinement demands sophisticated reasoning capabilities that only larger models can provide.
(2) \emph{Static Resource Allocation}: Most existing methods lack adaptive mechanisms to match computational resources to search phase complexity \cite{Chen2023EvoPromptingLM} \cite{guo2023connecting} , leading to either over-provisioning (wasted computation) or under-provisioning (poor search quality).
(3) \emph{Ambiguous Architecture Representation}: Many previous methods employ rigid or specially-tokenized architecture descriptions, which introduce ambiguity and comprehension challenges for LLMs, leading to frequent generation failures during the search process  \cite{zheng2023gpt4nas}\cite{wu2024evolutionary}.

These issues are particularly pronounced in complex downstream tasks such as object detection, where NAS must optimize not only accuracy but also constraints on parameters, FLOPs, and latency. Prior work like EfficientDet \cite{tan2019efficientdet} and NAS-FPN \cite{li2020nasfpn} shows the promise of architecture search in detection, but they often rely on hand-crafted search spaces and fixed search strategies.

To address these challenges, we propose PhaseNAS—a dynamic, LLM-based neural architecture search framework tailored for both lightweight classification tasks and complex perception tasks common in unmanned systems, such as object detection. Using YOLOv8~\cite{yolov8_ultralytics} as a baseline, we demonstrate how LLMs can dynamically generate and refine architectures that outperform hand-designed baselines in detection performance while remaining resource-efficient. Our method builds on recent LLM-based program synthesis efforts~\cite{rahman2024lemo,austin2021program}, incorporating structured templates to bridge natural language prompts with executable network components.

Unlike prior works that focus mainly on image classification, our framework extends LLM-based NAS to object detection tasks, highlighting its versatility in more complex vision scenarios. Specifically, \textbf{PhaseNAS} adopts a dynamic, two-phase LLM-based NAS: an initial broad exploration phase to cover diverse architectural candidates, followed by a refinement phase that iteratively improves promising designs. Phase transitions are guided by real-time score thresholds, ensuring efficient allocation of search resources and adaptive balance between exploration and exploitation.

Our work advances neural architecture search through the following key contributions:
\begin{enumerate}
    \item \textbf{Phase-Aware Dynamic Search}: We introduce a phase-aware controller that adaptively alternates between broad exploration and focused refinement, guided by the distribution of top-performing architectures. This eliminates manual scheduling and improves convergence efficiency for both image classification and object detection tasks.
    \item \textbf{Unified Architecture Template Language}: We propose a structured, parameterized template system that bridges natural language prompts and executable network code. This reduces semantic ambiguity and decoding failures, enabling robust, large-scale LLM-driven architecture generation for complex models such as YOLO-based detectors.
    \item \textbf{Detection-Specific Architecture Scoring}: We design a novel principled scoring approach for object detection architectures without full training, leveraging a Zen-Score-inspired metric tailored for detection. This enables rapid, resource-efficient architecture search in detection scenarios, which is previously unexplored.
    \item \textbf{Comprehensive Validation}: Extensive experiments on CIFAR-10/100 and COCO demonstrate the effectiveness of PhaseNAS. Our method discovers YOLOv8 variants with higher mAP and lower computation, and achieves up to 97.34\% accuracy on classification tasks with substantially reduced search costs.
\end{enumerate}

\section{Proposed Method}

The fundamental insight of PhaseNAS is that neural architecture search exhibits phase-dependent computational requirements. Early exploration benefits from broad sampling across the search space, which smaller language models can efficiently handle. Later refinement requires sophisticated reasoning about architectural trade-offs and subtle optimizations, demanding the advanced capabilities of larger models.
This observation leads to our core principle: \textbf{dynamically match computational resources to search phase complexity}. Rather than using a fixed LLM throughout the search, PhaseNAS adaptively transitions between appropriately-sized models based on real-time assessment of search progress.

\begin{figure}[htbp]
        \centering
        \includegraphics[width=\linewidth]{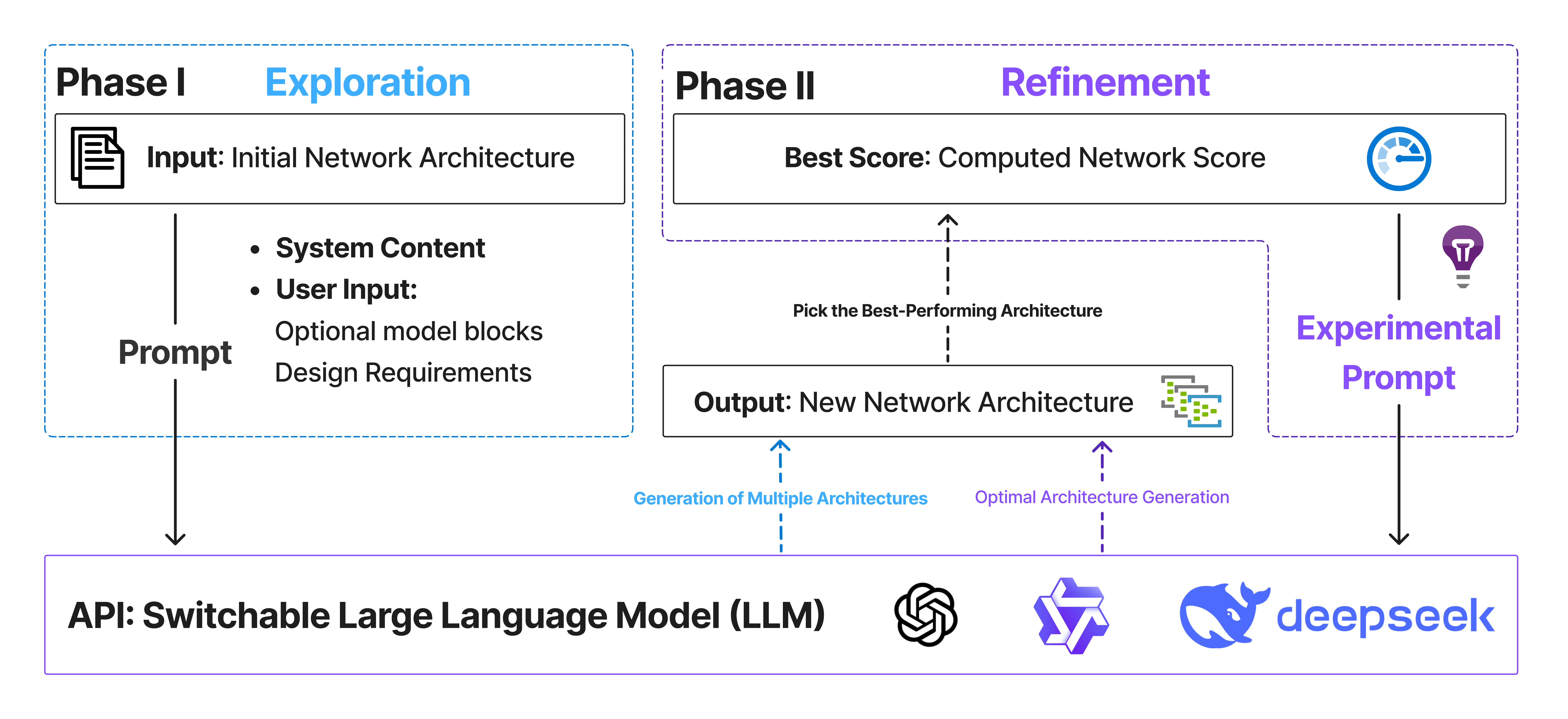}
        \caption{
        Overview of PhaseNAS: Phase I uses small LLM for efficient exploration, Phase II uses large LLM for focused refinement. The framework adaptively transitions between phases based on search progress.
        }
        \label{fig1}
\end{figure}

As illustrated in Figure \ref{fig1}, the algorithm's core innovation lies in its dynamic model scaling strategy, where the exploration phase utilizes a smaller, cost-efficient language model for broad and rapid architectural discovery, significantly reducing search cost. Once promising candidate architectures emerge, the refinement phase transitions to a larger, more capable language model, which further optimizes and diversifies these candidates with advanced reasoning and creativity. This two-stage design achieves an effective balance between search efficiency and the quality and novelty of the final architectures, governed by real-time score thresholds that control phase transitions and termination.  

Additionally, to support more advanced tasks beyond standard classification, PhaseNAS incorporates object detection through the YOLOv8 framework, which plays a critical role in enabling real-time perception and situational awareness in autonomous systems. This extension highlights the flexibility of our approach in adapting the search process to meet the stringent requirements of detection tasks, such as low latency, constrained FLOPs, and compact model size.

\subsection{Dynamic Search Process}

The search process operates within a constrained space $\mathbb{S}$ defined by three fundamental design principles:
\begin{enumerate}
    \item \textbf{Modular Component Selection}: All architectural components must be selected from predefined functional groups $\mathbb{G}_k$ (e.g., convolutional layers, residual blocks).
    \item \textbf{Dimensional Compatibility}: Adjacent modules must preserve dimensional consistency through strict channel matching, ensuring seamless integration.
    \item \textbf{Hardware-Aware Constraints}: The computational complexity of the generated architectures must remain within predefined limits, ensuring practical applicability across diverse deployment scenarios.
\end{enumerate}

The methodology progresses through two distinct phases, each designed to maximize efficiency and performance by leveraging the strengths of different LLM capacities.

\subsubsection{Exploration Phase}
In the exploration phase, a smaller and cost-efficient language model is employed to efficiently generate diverse architectural variants. The model receives natural language prompts encoding dimensional constraints, complexity boundaries, and task-specific requirements. The generated candidates are then evaluated by a scoring function and validated against resource constraints. Valid candidates are added to a quality-ordered pool, maintaining the top $K$ architectures.

\subsubsection{Refinement Phase}
Once any candidate in the pool achieves the transition threshold, the system transitions to the refinement phase. In this phase, a larger and more capable language model is used to further optimize and diversify the high-potential architectures based on performance feedback. The refinement process focuses on improving underperforming components while preserving the overall structure. The search terminates when the stopping threshold is reached, ensuring efficient progress through the search space.

\subsection{Search Space Definition}
The design of an effective NAS framework fundamentally depends on a well-defined search space for different tasks. In PhaseNAS, the search space is systematically tailored for both image classification and object detection to maximize efficiency, compatibility, and innovation.

\paragraph{Search Space for Classification Tasks}

For classification, the search space $\mathbb{S}_{\text{cls}}$ is made explicit and highly structured to facilitate efficient search and reduce semantic ambiguity. The core constraints are as follows:

\begin{enumerate}
    \item \textbf{Available Building Blocks}: Each network is constructed from a predefined set of residual and convolutional blocks with various kernel sizes and activation functions. The search is restricted to combinations of these standardized modules, ensuring compatibility and reproducibility across all candidate architectures.
    \item \textbf{Channel Compatibility}: The output channel of any block must match the input channel of the subsequent block to ensure seamless tensor propagation.
    \item \textbf{Input/Output Constraints}: The model input must be a three-channel image, and the output must be suitable for the classification head.
    \item \textbf{Resource Constraints}: The parameter count and network FLOPs are constrained within practical deployment ranges to ensure efficiency.
\end{enumerate}

This explicit, template-based design ensures that LLMs can reliably generate, interpret, and modify architectures with minimal risk of invalid structures or compilation errors.

\paragraph{Search Space for Object Detection Tasks}

For object detection, the search space $\mathbb{S}_{\text{det}}$ is designed to extend and enhance the YOLOv8 family of models while adhering to strict resource constraints. The main characteristics are:

\begin{enumerate}
    \item \textbf{YOLOv8 as Baseline}: The search is anchored on the official YOLOv8n and YOLOv8s backbones, allowing modifications and reordering of backbone, neck, and head blocks, while constraining overall parameters and FLOPs to remain similar to the base models.
    \item \textbf{Block-Level Modifications}: Candidate architectures are generated by reconfiguring, replacing, or inserting blocks within the backbone and neck, provided input/output channels and tensor shapes remain compatible.
    \item \textbf{Innovative Block Extensions}: 
    \begin{itemize}
        \item \textbf{YOLOv8+}: Models in this variant are generated by reorganizing existing YOLOv8 blocks and tuning their parameters, without introducing new block types.
        \item \textbf{YOLOv8*}: In addition to the modifications allowed in YOLOv8+, this variant explicitly introduces two novel blocks:
            \begin{itemize}
                \item \textbf{SCDown}\cite{wang2024yolov10}: A spatial compression downsampling module, designed to enhance multi-scale feature extraction and improve information flow in the early backbone.
                \item \textbf{PSA}\cite{wang2024yolov10}: A polarized self-attention block, which introduces channel-wise attention to boost target localization and feature selectivity.
            \end{itemize}
        These custom blocks are only available in the YOLOv8* search space, enabling the framework to explore more expressive and powerful architectures beyond the original YOLOv8 design.
    \end{itemize}
    \item \textbf{Multi-Scale and Detection Head Compatibility}: All searched architectures must support multi-scale feature outputs and remain compatible with the YOLO detection heads.
    \item \textbf{Resource Constraints}: Parameter count and FLOPs are kept within the original YOLOv8n/s budgets to ensure real-time performance and fair comparison.
\end{enumerate}

\textbf{Summary:} The classification search space is strictly limited to ten modular blocks with clear parameter and depth constraints. The detection search space, while based on YOLOv8, is extended in YOLOv8* by the inclusion of two advanced modules (SCDown and PSA), providing greater architectural diversity and enabling higher mAP under resource constraints.

\subsection{Algorithm Overview}
Algorithm~\ref{alg:phase_nas} outlines the full PhaseNAS search procedure, dynamically alternating between LLM-driven exploration and efficient refinement based on real-time score thresholds.

In our implementation, we define $\mathcal{M}_E$ with multiple variants, including Qwen2.5-7B, Qwen2.5-14B, and Qwen2.5-32B (7-32 billion parameters) for cost-efficient exploration, and $\mathcal{M}_R$ as Qwen2.5-72B, Llama-3.3-70B and Claude-3.5-Sonnet (70+ billion parameters) for high-quality refinement. The choice of model sizes can be adapted based on available computational resources, with the key principle being $|\mathcal{M}_E| < |\mathcal{M}_R|$ to ensure progressive capability scaling.

\begin{algorithm}[ht]
\caption{PhaseNAS Architecture Search}
\label{alg:phase_nas}
\begin{algorithmic}[1]

\Require Initial architecture $\mathcal{S}_{\text{init}}$, small LLM $\mathcal{M}_E$ (for Exploration), large LLM $\mathcal{M}_R$ (for Refinement), thresholds $\gamma_{\text{trans}}, \gamma_{\text{stop}}$, pool size $K$

\Ensure Optimal architecture $\mathcal{S}^*$

\State \textbf{Initialize:} $\mathbb{A}_c \gets \{\mathcal{S}_{\text{init}}\}$, $\phi \gets$ Exploration Phase

\While{$\max_{\mathcal{S} \in \mathbb{A}_c} \mathcal{E}_z(\mathcal{S}) < \gamma_{\text{stop}}$}
    \If{$\phi = $ Exploration}
        \State Generate $\mathcal{S}_{\text{new}} \gets \mathcal{M}_E(\text{exploration\_prompt})$ \Comment{Small LLM for cost-efficient exploration}

        \If{$\mathcal{V}(\mathcal{S}_{\text{new}})$ and $\mathcal{E}_z(\mathcal{S}_{\text{new}}) > \min_{\mathcal{S} \in \mathbb{A}_c} \mathcal{E}_z(\mathcal{S})$}
            \State Add $\mathcal{S}_{\text{new}}$ to $\mathbb{A}_c$
            \If{$|\mathbb{A}_c| > K$}
                \State Remove architecture with lowest $\mathcal{E}_z$ from $\mathbb{A}_c$
            \EndIf
        \EndIf
        \If{$\exists \mathcal{S} \in \mathbb{A}_c$ such that $\mathcal{E}_z(\mathcal{S}) \geq \gamma_{\text{trans}}$}
            \State $\phi \gets$ Refinement Phase
            \State Set base architecture $\mathcal{S}_{\text{base}} \gets \arg\max_{\mathcal{S} \in \mathbb{A}_c} \mathcal{E}_z(\mathcal{S})$
        \EndIf
    \Else \Comment{Refinement Phase}
        \State Generate refined architecture $\mathcal{S}_{\text{new}} \gets \mathcal{M}_R(\mathcal{S}_{\text{base}}, \text{feedback})$ \Comment{Large LLM for high-quality refinement}
        \If{$\mathcal{V}(\mathcal{S}_{\text{new}})$ and $\mathcal{E}_z(\mathcal{S}_{\text{new}}) > \mathcal{E}_z(\mathcal{S}_{\text{base}})$}
            \State Update base architecture: $\mathcal{S}_{\text{base}} \gets \mathcal{S}_{\text{new}}$
            \State Add $\mathcal{S}_{\text{new}}$ to $\mathbb{A}_c$
        \EndIf
    \EndIf
\EndWhile
\State \Return $\mathcal{S}^* \gets \arg\max_{\mathcal{S} \in \mathbb{A}_c} \mathcal{E}_z(\mathcal{S})$
\end{algorithmic}
\end{algorithm}

\subsection{LLM-Compatible Architecture Representation}
A critical challenge in LLM-based NAS is the semantic mismatch between natural language architecture descriptions and executable implementations \cite{wu2024evolutionary}. To address this, PhaseNAS adopts a structured template language inspired by the principles outlined in \cite{dong2020bench}. For example:
A convolutional block is represented as \texttt{ConvK3BNRELU(3,8,1,1)}, which specifies the kernel size, the input/output channels, and the stride.
Residual blocks use parameterized templates such as \texttt{ResK3K3(16,32,2,1)}, ensuring consistency across layers.

This structured representation serves as an interpretable interface between LLMs and compilers, reducing compilation errors and improving the success rate of architecture evaluation. Moreover, it facilitates seamless integration with frameworks like YOLOv8, where block-based modularity is essential for designing efficient object detection architectures.

\subsection{Task Adaptation: From Classification to Object Detection}

To guide the architecture search process effectively, PhaseNAS employs task-specific NAS score computation methods for evaluating candidate architectures. These scores are designed to quantify the architecture's response to input perturbations and its normalization stability, offering fast proxies for generalization without full training. Below, we present the scoring procedures for classification and object detection tasks, including the adaptations made for the YOLO framework.

\subsubsection{NAS Score for Classification}
\label{sec:nas-score-class}
We follow the scoring function proposed in Zen-NAS~\cite{lin2021zen},
which effectively captures the model's sensitivity to perturbation and the stability of Batch Normalization layers. As these calculations remain the same as in Zen-NAS, we omit the detailed formula derivations here for brevity.

\subsubsection{NAS Score for Object Detection}
\label{sec:nas-score-detect}
Object detection tasks require feature pyramids and larger input sizes. Thus, the NAS score is extended to measure the consistency of extracted multi-scale feature maps under input perturbations.

\paragraph{1. Input Perturbation}

\begin{equation}
    x_{\text{mix}} = x_1 + \gamma \cdot x_2.
\end{equation}

\paragraph{2. Feature Map Extraction}

Given model \( M(\cdot) \), let \( F(\cdot) \) extract the internal feature maps. We compute:
\begin{equation}
    f_1 = F(M(x_1)), \quad f_{\text{mix}} = F(M(x_{\text{mix}})),
\end{equation}
with \( f^{(l)} \in \mathbb{R}^{b \times c_l \times h_l \times w_l} \) as the feature map at scale \( l \).

\paragraph{3. Multi-Scale Difference}

Total feature perturbation response is:
\begin{equation}
    \Delta = \sum_{l=1}^{L} \|f^{(l)}_1 - f^{(l)}_{\text{mix}}\|_1,
\end{equation}
where \( L \) is the number of feature scales.

\paragraph{4. BatchNorm Scaling}

\begin{equation}
    \mathcal{B} = \sum_{m} \log\left( \sqrt{\text{running\_var}_m + \epsilon} \right).
\end{equation}

\paragraph{5. Final NAS Score}

The score for detection is defined as:
\begin{equation}
    \text{NAS}_{\text{det}} = \log(\Delta + \epsilon) + \mathcal{B}.
\end{equation}

\paragraph{6. Repetition and Aggregation}

Repeat the process \( R \) times and compute:
\begin{equation}
    \mu = \frac{1}{R} \sum_{i=1}^{R} s_i, \quad \sigma = \sqrt{ \frac{1}{R} \sum_{i=1}^{R} (s_i - \mu)^2 },
\end{equation}
where \( \mu \) is used for ranking and selection.

\vspace{1em}

\noindent
The NAS scores for classification and detection provide a unified yet adaptable framework for evaluating candidate architectures efficiently. By extending the scoring mechanism to handle multi-scale outputs and detector-specific challenges, PhaseNAS ensures robust and scalable neural architecture search across a wide range of AI tasks.


\section{Experimental Evaluation}
\paragraph{Experimental Design Rationale.} 
Our experiments are crafted to verify that PhaseNAS simultaneously lowers search cost via dynamic LLM scaling, discovers architectures that match or surpass state-of-the-art NAS baselines, and generalizes across both classification and detection tasks within a single resource-adaptive framework.

We first evaluate the effectiveness of PhaseNAS by directly comparing it with recent LLM-based NAS baselines, specifically GUNIUS, on the NAS-Bench-Macro dataset. This not only demonstrates the performance advantage of our dynamic, phase-aware approach but also validates the benefit of adaptive model switching. We then extend our study to both image classification and object detection domains, applying PhaseNAS to multiple search spaces and tasks to showcase its scalability and general applicability.

\subsection{Comparison with LLM-based NAS Baselines}

\begin{figure*}[ht]
  \centering
  \begin{subfigure}{1\linewidth}
    \includegraphics[width=\linewidth]{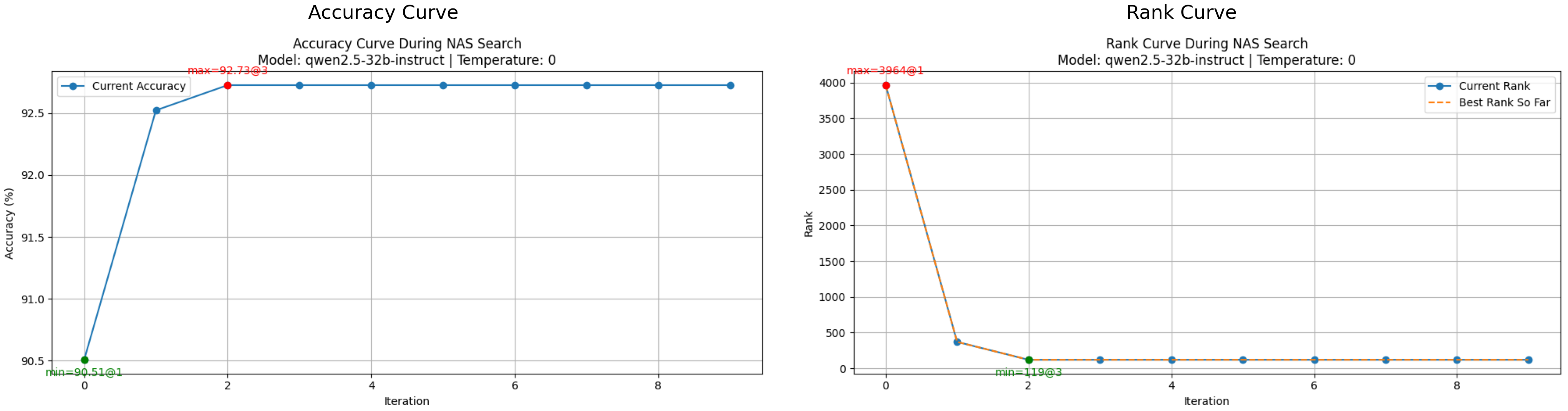}
    \caption{GUNIUS (Qwen2.5-32B)}
  \end{subfigure}

  \begin{subfigure}{1\linewidth}
    \includegraphics[width=\linewidth]{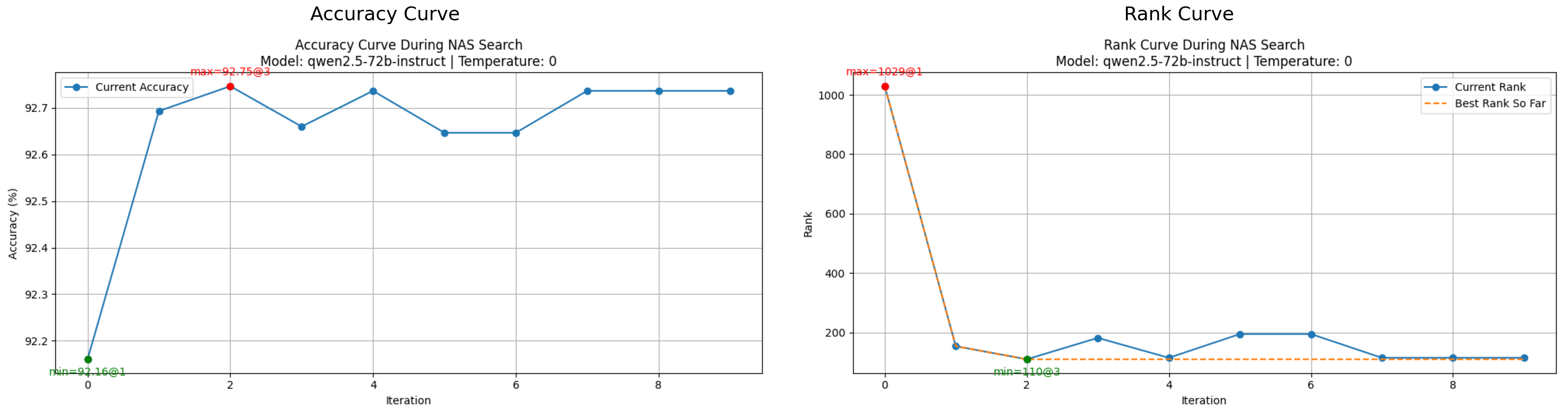}
    \caption{GUNIUS (Qwen2.5-72B)}
  \end{subfigure}

  \begin{subfigure}{1\linewidth}
    \includegraphics[width=\linewidth]{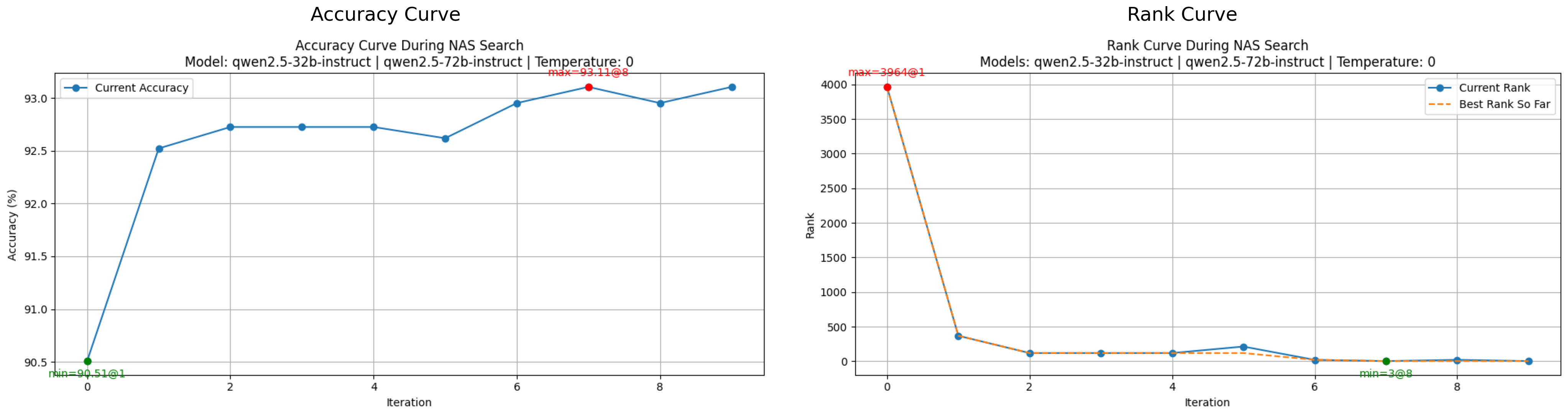}
    \caption{PhaseNAS (Qwen2.5-32B $\rightarrow$ 72B)}
  \end{subfigure}
  \caption{Accuracy and rank curves on NAS-Bench-Macro for three approaches.}
\label{fig:llm_nas_comparison}
\end{figure*}

To demonstrate the core effectiveness of PhaseNAS, we first benchmark it against the recent LLM-based NAS method GUNIUS on the NAS-Bench-Macro dataset. GUNIUS utilizes Qwen2.5-32B and Qwen2.5-72B as independent architecture generators, while our PhaseNAS dynamically switches between these two models in different search phases.

\paragraph{Experimental Setup.}
All methods are evaluated on the NAS-Bench-Macro dataset under identical search protocols. The original GUNIUS paper adopts GPT-4 as the LLM backbone; for resource and accessibility reasons, we instead use Qwen2.5-32B and Qwen2.5-72B as LLM generators for both GUNIUS and our PhaseNAS. For GUNIUS, each model is tested independently as the architecture generator. For PhaseNAS, the phase-aware controller adaptively alternates between Qwen2.5-32B and Qwen2.5-72B based on candidate performance distribution. Each search run is repeated for 10 iterations to ensure convergence.

\paragraph{Results and Analysis.}
As shown in Fig.~\ref{fig:llm_nas_comparison}, all three approaches are evaluated under consistent search protocols. For GUNIUS with Qwen2.5-32B, the best architecture achieves an accuracy of 92.73\% with a final rank of 119. When using Qwen2.5-72B, GUNIUS yields a slightly better result: the top architecture reaches 92.75\% accuracy with a rank of 110. In contrast, our PhaseNAS approach achieves a significantly higher accuracy of 93.11\% and a much lower (better) rank of 3, indicating that it not only discovers more accurate architectures but also identifies solutions that are closer to the global optimum in the search space.

This comparison illustrates that PhaseNAS's phase-aware dynamic search and adaptive model switching are substantially more effective than using a single LLM in isolation. The consistent improvement in both accuracy and rank across the search process demonstrates the practical advantage and superiority of our method.

\subsection{Generalization to Classification and Detection Tasks}
Building upon the LLM-based NAS benchmark results, we further evaluate PhaseNAS on both image classification and object detection tasks to verify its generalizability and scalability. The classification experiments are conducted on CIFAR-10 and CIFAR-100, while object detection experiments are performed on the COCO validation set. To ensure fair and meaningful architecture comparisons, we impose the following consistent constraints during the search process:

\begin{itemize}
    \item \textbf{Model Size}: Limit total parameters for edge device deployment.
    \item \textbf{FLOPs}: Bound computational cost for real-time efficiency.
    \item \textbf{Latency}: Ensure practical inference speed on target hardware.
    \item \textbf{Depth}: Prevent over-deep, hard-to-optimize models.
\end{itemize}
All methods are evaluated using the same NAS scoring functions, and all training pipelines are aligned across baselines for fairness.

\subsection{Classification Results and Analysis}
We compare the efficiency and performance of \textbf{PhaseNAS} and the widely-used \textbf{Zen-NAS} on CIFAR-10 and CIFAR-100, ensuring a fair comparison by adopting the same Zen-score evaluation. Table~\ref{tab1} reports the results for different search settings.

\begin{table}[ht]
\centering
\begin{tabular}{|c|c|c|c|c|}
\hline
\textbf{Method} & \textbf{Zen-Score} & \textbf{Search Time (min)} & \textbf{CIFAR-10 Acc.} & \textbf{CIFAR-100 Acc.} \\
\hline
Zen-NAS & 99.20 & 7.43 & 96.76 & 79.98 \\
PhaseNAS & \textbf{99.43} & \textbf{1.20} & 96.63 & \textbf{80.74} \\
\hline
Zen-NAS & 111.63 & 19.07 & 96.02 & 80.81 \\
PhaseNAS & \textbf{111.33} & \textbf{7.06} & \textbf{96.63} & \textbf{81.24} \\
\hline
Zen-NAS & 121.38 & 67.46 & 96.96 & 81.26 \\
PhaseNAS & \textbf{121.44} & \textbf{9.01} & \textbf{97.34} & \textbf{81.36} \\
\hline
\end{tabular}
\caption{Comparison between \textbf{PhaseNAS} and \textbf{Zen-NAS} on classification benchmarks. Higher Zen-scores and lower search times indicate better efficiency.}
\label{tab1}
\end{table}

As shown in Table~\ref{tab1}, \textbf{PhaseNAS} achieves up to \textbf{86\% reduction} in search time compared to Zen-NAS, comparable or improved classification accuracy (especially on CIFAR-100), and comparable Zen-scores, indicating better architectural potential.

These results demonstrate that PhaseNAS can significantly accelerate the NAS process while maintaining or improving final accuracy. This efficiency gain is attributed to the dynamic use of large and small language models and the phase-aware search control, as verified earlier in the LLM-based NAS macro-benchmark.

\subsection{Object Detection Results and Analysis}
To further validate the versatility of PhaseNAS, we apply it to object detection on COCO using the YOLOv8 framework. Architecture search is conducted under the same FLOPs and parameter constraints as YOLOv8n and YOLOv8s, with LLM-guided prompts driving both exploration and refinement.

\begin{table}[ht]
\centering
\small
\begin{tabular}{|c|c|c|c|c|}
\hline
\textbf{Family} & \textbf{Model} & \textbf{mAP@50:95} & \textbf{Params (M)} & \textbf{FLOPs (G)} \\
\hline
\multirow{3}{*}{YOLOv8n} 
& YOLOv8n & 37.3 & 3.2 & 8.7 \\
& YOLOv8n+ & \textbf{38.0} & 3.0 & 8.2 \\
& YOLOv8n* & \textbf{39.1} & 2.95 & 8.5 \\
\hline
\multirow{3}{*}{YOLOv8s} 
& YOLOv8s & 44.9 & 11.2 & 28.6 \\
& YOLOv8s+ & \textbf{45.4} & 10.3 & 25.0 \\
& YOLOv8s* & \textbf{46.1} & 9.9 & 22.4 \\
\hline
\end{tabular}
\caption{
Architecture quality comparison of YOLO variants generated by PhaseNAS on COCO validation set. All PhaseNAS variants achieve higher mAP with reduced computational cost, demonstrating effective architectural optimization beyond manual design.
(+) indicates models optimized within original block constraints; (*) indicates models with novel architectural components.
}
\label{tab:yolov8_compare}
\end{table}
Table~\ref{tab:yolov8_compare} shows that PhaseNAS generates detection models that not only outperform the hand-designed YOLOv8 baselines in mAP, but also reduce model size and computational complexity:
\begin{enumerate}
    \item \textbf{YOLOv8n series}: YOLOv8n+ and YOLOv8n* improve mAP to 38.0 and 39.1 respectively, both with fewer parameters and lower FLOPs than the original (37.3, 3.2M, 8.7G).
    \item \textbf{YOLOv8s series}: YOLOv8s+ and YOLOv8s* achieve mAPs of 45.4 and 46.1 respectively, while reducing parameter count and FLOPs compared to the baseline (44.9, 11.2M, 28.6G).
\end{enumerate}

\paragraph{Analysis of Results.}
The superior performance of PhaseNAS can be attributed to three key factors: (1) \textbf{Adaptive resource allocation}: The dynamic phase transition allows efficient exploration with small LLMs followed by focused refinement with large LLMs, reducing overall computational cost while maintaining search quality. (2) \textbf{Structured representation}: The template-based architecture encoding reduces semantic ambiguity and improves LLM understanding compared to free-form descriptions. (3) \textbf{Task-specific scoring}: The extended NAS score for detection tasks captures multi-scale feature quality, which is crucial for object detection performance.

\subsection{Summary of Results}
From NAS-Bench-Macro to real-world tasks, PhaseNAS consistently demonstrates:
\begin{itemize}
    \item Superior search efficiency with substantially reduced computation time,
    \item Competitive or improved accuracy under tight deployment constraints,
    \item Applicability to both classification and detection tasks,
    \item Effective progressive search leveraging both large and small language models.
\end{itemize}
These results collectively validate PhaseNAS as a flexible, scalable, and practical NAS framework for a wide range of AI applications.


\section{Conclusion}
We present \textbf{PhaseNAS}, a dynamic NAS framework that adaptively balances exploration and refinement through phase transitions guided by real-time score thresholds and dynamic selection of large language models (LLMs) with different capacities. 
Our experiments show that PhaseNAS consistently surpasses state-of-the-art LLM-based NAS methods such as GUNIUS on the NAS-Bench-Macro benchmark, discovering architectures with higher accuracy and achieving much better rankings in the search space through its phase-aware search and adaptive model switching.
For image classification, PhaseNAS reduces search time by up to 86\% compared to Zen-NAS, while maintaining or even improving top-1 accuracy on CIFAR-10 and CIFAR-100. In object detection, PhaseNAS automatically produces YOLOv8 variants that outperform baseline models in both mAP and computational efficiency, all without manual architecture tuning.
These results demonstrate the versatility and efficiency of PhaseNAS as a general NAS solution for diverse vision tasks. In future work, we plan to extend PhaseNAS to broader domains such as natural language processing and multimodal applications, integrate more hardware-aware metrics (e.g., latency, energy consumption), and further enhance adaptive phase transitions for real-world AI deployment.

\end{document}